\documentclass{article}

\usepackage[final]{nips_2018}

\usepackage[utf8]{inputenc} % allow utf-8 input
\usepackage[T1]{fontenc}    % use 8-bit T1 fonts
\usepackage{hyperref}       % hyperlinks
\usepackage{url}            % simple URL typesetting
\usepackage{booktabs}       % professional-quality tables
\usepackage{amsfonts}       % blackboard math symbols
\usepackage{nicefrac}       % compact symbols for 1/2, etc.
\usepackage{microtype}      % microtypography
\usepackage{times}  %Required
\usepackage{helvet}  %Required
\usepackage{courier}  %Required
\usepackage{url}  %Required
\usepackage{graphicx, xcolor}  %Required
\usepackage{bm}
\usepackage{amsmath}
\usepackage{amsfonts}
\usepackage{multirow} %Yuhui
\usepackage{longtable} %Yuhui
\usepackage{supertabular} %Yuhui
\usepackage{colortbl} %Yuhui
\usepackage{float} %Yuhui
\usepackage{booktabs} %Yuhui
\def\reals{\mathbb{R}}
\newcommand{\capital}[1]{\bm{\mathrm{#1}}}
\renewcommand{\exp}[1]{\operatorname{exp}\left(#1\right)} %

\title{Large-scale Generative Modeling to Improve Automated Veterinary Disease Coding}  % Language Modeling

\author{
  Yuhui Zhang$^{1,+}$, Allen Nie$^{2,+}$, James Zou$^{2,*}$ \\
  $^{1}$Department of Computer Science and Technology, Tsinghua University, Beijing, China \\
  $^{2}$Department of Biomedical Data Science, Stanford University, Stanford, CA 94305, USA \\
  $^{+}$equal contribution, $^{*}$jamesz@stanford.edu. % I can not insert any information due to space restriction....
} 

\begin{document}

% [] Reviewer1: a statistical test is preferred to determine the significance of improvements. 
% [X] Reviewer2: it would be interesting to see variation of data characteristics beyond the counts provided (e.g. disease types, species, etc)
% [X](Explain in Appendix) Reviewer2: Details on how the datasets can be obtained should be provided for reproducibility purposes 
% [X](Explain in Appendix) Reviewer2: provide a clearer description on how datasets were split for training/testing.

\maketitle

\begin{abstract}
Supervised learning is limited both by the quantity and quality of the labeled data. In the field of medical record tagging, writing styles between hospitals vary drastically. The knowledge learned from one hospital might not transfer well to another. This problem is amplified in veterinary medicine domain because veterinary clinics rarely apply medical codes to their records. We proposed and trained the first large-scale generative modeling algorithm in automated disease coding. We demonstrate that generative modeling can learn discriminative features when additionally trained with supervised fine-tuning. We systematically ablate and evaluate the effect of generative modeling on the final system's performance. We compare the performance of our model with several baselines in a challenging cross-hospital setting with substantial domain shift. We outperform competitive baselines by a large margin. In addition, we provide interpretation for what is learned by our model.
\end{abstract}

\section{Introduction}

One of the most significant challenges for veterinary data science is that veterinary primary practices rarely code clinical findings in EHR records. This makes it hard to perform core tasks like case finding, cohort selection, or to support the production of basic descriptive statistics like disease prevalence. It is becoming increasingly accepted that spontaneous diseases in animals have important translational impact on the study of human disease for a variety of disciplines \citep{kol2015companion}. Beyond the study of zoonotic diseases, which represent 60-70\% of all emerging diseases, non-infectious diseases, like cancer, have become increasingly studied in companion animals as a way to mitigate some of the problems with rodent models of disease \citep{leblanc2016defining}. Additionally, spontaneous models of disease in companion animals are being used in drug development pipelines as these models more closely resemble the ``real world'' clinical settings of diseases than genetically altered mouse models~\citep{grimm2016bark,klinck2017translational,baraban2014new,hernandez2018naturally}. 

% TODO: need change!!!!!
In comparison to the human EHR, there has been little ML work on veterinary EHR, which faces a unique challenge. The labeled data which are accessible to research only reside in referral teaching hospitals. These hospitals often specialize in a specific type of diseases. The patient type, as well as the disease distributions, do not resemble the general population. Machine learning models trained on this dataset might easily get biased and perform poorly on general clinical records. We refer to this as the \textbf{cross-hospital challenge}.

\paragraph{Our contributions} 
We develop an algorithm to leverage one million unlabeled clinical notes through generative sequence modeling, and demonstrate such large-scale modeling can substantially improve the model's performance in a cross-hospital setting.
We adapt the new state-of-the-art Transformer model proposed by \citet{vaswani2017attention}.
We systematically evaluate the model performance in this cross-hospital setting, where the algorithm trained on one hospital is evaluated in a different hospital with substantial domain shift. In addition, we provide interpretation for what is learned by the deep network. Our algorithm addresses an important application in healthcare, and our experiments add insights into the power of generative sequence modeling for clinical NLP.  

\section{Task and Data}

We formulate the problem of automated disease coding as a multi-label classification problem. Given a veterinary record $\capital{X}$, which contains detailed description of the diagnosis, we try to infer a subset of diseases $\bm{y} \in \mathcal{\bm{Y}}$, given a pre-defined set of diseases $\mathcal{\bm{Y}}$. The problem of inferring a subset of disease codes can be viewed as a series of independent binary prediction problems~\citep{sorower2010literature}. 

We use three datasets in this work (Appendix Table~\ref{table:pdtb-implicit}). 
\textbf{CSU(Labeled)}: We use a curated set of 112,558 veterinary notes from the Colorado State University College of Veterinary Medicine and Biomedical Sciences. Each note is annotated with a set of SNOMED-CT codes by veterinarians at Colorado State. \textbf{PP(Labeled)}: We obtain a smaller set of 586 discharge summaries curated from a commercial veterinary practice located in Northern California. Two veterinary experts applied SNOMED-CT codes to these records and achieved consensus on the records used for validation. This dataset is drastically different from the CSU dataset evidenced by their shorter length and usage of abbreviations. \textbf{SAGE(Unlabeled)}: We obtained a large set of 1,019,747 unlabeled notes from the SAGE Centers for Veterinary Specialty \& Emergency Care. This is a set of raw clinical notes without any codes applied to them. The characteristics of this dataset should be similar to the PP dataset because they are both primary local clinics.

\begin{figure}[t!]
    \centering
    \includegraphics[width=4.8in]{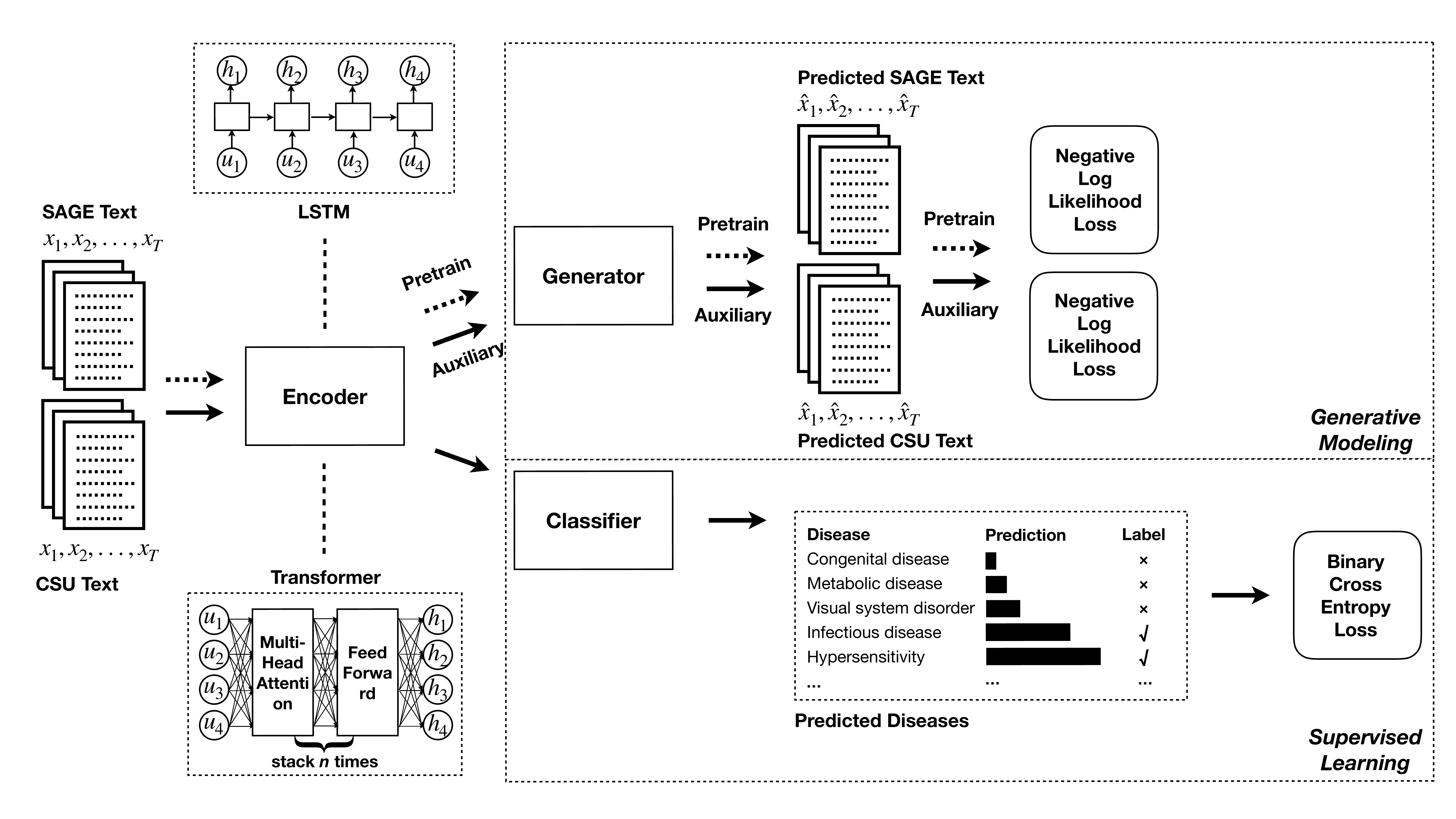}
    \caption{Our proposed model architecture for automated disease coding. Two tasks are shown: generative modeling (top) and supervised learning (bottom).  The dashed arrows represent the generative modeling process on the unlabeled SAGE data, and the solid arrows represent the supervised learning process on the labeled CSU data. An additional test is done on the PP data (not shown). }
    \label{fig:model}
\end{figure}

\begin{table}[t!]
\centering
\begin{tabular}{l|cccc|cccc}
\toprule
 & \multicolumn{4}{c|}{CSU} & \multicolumn{4}{c}{PP (Cross-hospital)} \\
Model & EM & P & R & $F_1$ & EM & P & R & $F_1$ \\
\midrule
Metamap(SVM) & 32.2 & 74.8 & 75.0 & 74.8 & 3.2 & 57.3 & 53.1 & 51.6\\
Metamap(MLP) & 41.2 & 82.6 & 71.8 & 76.4 & 13.8 & 56.4 & 47.6 & 50.5\\
CAML\citep{mullenbach2018explainable} & 46.7 & 86.9 & 76.1 & 80.5 & 16.9 & 72.2 & 50.2 & 54.7 \\
\midrule
LSTM & 46.7 & 87.5 & 74.2 & 79.5 & 17.8 & 74.8 & 49.6 & 54.9 \\
% LSTM+MetaMap & 46.0 & 86.9 & 74.8 & 79.7 & 16.8 & 75.8 & 47.9 & 52.7 \\
LSTM+Word2Vec & 47.2 & 86.4 & 76.6 & 80.7 & 20.4 & 75.8 & 48.6 & 54.3 \\
LSTM+Pretrain & 48.2 & 87.4 & 76.2 & 81.0 & 20.1 & 73.8 & 52.0 & 57.6 \\
LSTM+Auxiliary & 49.2 & 88.2 & 76.0 & 81.0 & 20.8 & 75.2 & 53.8 & 58.7 \\
LSTM+Auxiliary+Pretrain & 49.0 & 87.6 & 76.5 & 81.2 & 19.6 & 75.5 & 54.8 & 60.3 \\
\midrule
Transformer & 45.1 & 86.3 & 73.5 & 78.6 & 17.3 & 73.3 & 54.5 & 58.5 \\
% Transformer+MetaMap & 44.8 & 87.7 & 72.8 & 78.6 & 14.3 & 72.6 & 50.5 & 54.8 \\
Transformer+Word2Vec & 41.2 & 87.2 & 68.9 & 75.2 & 18.8 & 77.1 & 50.5 & 55.2 \\
Transformer+Pretrain & 46.6 & 87.5 & 74.6 & 79.6 & 19.9 & 73.5 & 50.5 & 55.4 \\
Transformer+Auxiliary & 49.4 & 87.3 & 78.5 & 82.2 & 22.2 & 75.0 & 59.3 & 63.6 \\
\textbf{Transformer+Auxiliary+Pretrain} & \textbf{50.1} & \textbf{88.3} & \textbf{77.4} & \textbf{81.8} & \textbf{25.2} & \textbf{75.4} & \textbf{64.5} & \textbf{68.0} \\
\bottomrule
\end{tabular}
\caption{Evaluation of trained classifiers on the CSU test data and PP data. \textbf{EM} is the fraction of cases where the set of diseases predicted by the model \emph{exactly matches} the expert labels. The classifiers are trained on a subset of CSU. Notation: \textbf{LSTM} and \textbf{Transformer} are our two base encoder models; \textbf{+Word2Vec} uses Word2Vec trained on SAGE to initialize; \textbf{+Pretrain} uses generative modeling loss $-\log p(X)$ on SAGE to initialize; \textbf{+Auxiliary} uses generative modeling loss on CSU in addition to classification objective on CSU: $L(\mathcal{C})- \lambda * \log p(X)$. }
\label{tab:overall-csu-pp}
\end{table}

\section{Our Model}

Our proposed model architecture is shown in Figure~\ref{fig:model}. Two tasks are shown: generative modeling and supervised learning. We describe these two tasks in the following section. 

\subsection{Generative Modeling}

A generative model over text is also referred to as a language model. Text sequence is an ordered list of tokens. Therefore, we can build an autoregressive model to estimate the joint probability of the entire sequence: $p(X) = p(x_1, ..., x_T)$. In an ordered sequence, we can factorize it as $p(X) = \prod_{t=1}^T p(x_t|x_1, ..., x_{t-1})$. Concretely, we estimate the token distribution of $x_t$ by using the contextualized representation provided by our encoder: $h_t = \mathrm{Encoder}(h_1, ..., h_{t-1})$. We optimize over the negative log-likelihood of the distribution $-\log p(X) = - \sum_{t=1}^T \log p(x_t|x_1,...,x_{t-1})$.
% The benefit of generative modeling is that since we are modeling the data distribution $p(X)$,  % <Yuhui> ??

In our model, we examine the effect of generative modeling on two encoder architectures: Transformer and the Long Short-Term Memory (LSTM)~\citep{hochreiter1997long}. We use this objective in two parts of our system: 1) \textbf{pretrain} encoder's parameters; 2) serve as an \textbf{auxiliary task} during training of the classifier. 

\subsection{Supervised Learning}

Classifier uses a dot-product attention layer to get a summary representation $c$ for the entire sequence. We describe the computation in Appendix Eqn~\ref{eq:projection-layer1}. We then use a fully connected layer to down project it and calculate probability: $p(y_j) = \sigma(w_{j}^T c + b_{j})$. We compute the binary cross entropy loss across $m$ labels: $L(\mathcal{C}) = - \frac{1}{m} \sum_{j=1}^m y_j \log p(y_j) + (1-y_j) \log (1-p(y_j))$.

Finally, we use a mixture of two losses $L_{\mathrm{total}} = L(\mathcal{C}) - \lambda * \log p(X)$ and use hyperparameter $\lambda=0.5$ to set the strength of the auxiliary task loss when we use generative modeling as an \textbf{auxiliary task} in our classification training. 

\begin{table}[t!]
\centering
\begin{tabular}{l|l}
\toprule
Disease (SNOMED-CT code) & Extracted Keywords \\
\midrule
\multirow{2}{*}{Traumatic AND/OR non-traumatic injury} & fracture, wound, laceration, due, assessment, \\
 & trauma, this, bandage, time, owner \\
\midrule
\multirow{2}{*}{Visual system disorder} & eye, ophthalmology, surgery, eyelid, assessment, \\
 & sicca, time, uveitis, diagnosed, this \\
\midrule
\multirow{2}{*}{Hypersensitivity condition} & dermatitis, allergic, therapy, atopic, otitis, \\
 & pruritus, ears, assessment, allergies, dermatology \\
\midrule
\multirow{2}{*}{Metabolic disease} & diabetes, nph, hypercalcemia, glargine, vetsulin, \\
 & weeks, home, insulin, amlodipine, dose \\
\midrule
\multirow{2}{*}{Anemia} & pancytopenia, anemia, visit, hemolytic, persistent, \\
 & steroids, hypertension, neoplasia, exam, thickening \\
\bottomrule
\end{tabular}
\caption{Most influential words in the best model (Transformer+Auxiliary+Pretrain). We select five representative disease categories. For each disease, we show the top 10 words in the MetaMap medical dictionary that the model most strongly associates with the disease. }
\label{tab:topwords}
\end{table}

\section{Results}

We conduct systematic experiments on different models and ablations to quantify which component of our model improves the automatic coding performance (Table~\ref{tab:overall-csu-pp}).

\paragraph{Neural networks outperform feature-based models}

We use the popular MetaMap, a program developed by the National Library of Medicine (NLM)~\citep{aronson2010overview}, as a baseline. MetaMap processes a document and outputs a list of matched medically-relevant keywords with its frequencies in the given document. 
We directly train on the sparse bag-of-words feature representation from MetaMap. We use SVM or MLP as the classification algorithm from scikit-learn \citep{scikitlearn}.
We find its performance is worse than the CAML, LSTM and Transformer on both the CSU and PP test data.  

\paragraph{Generative modeling outperforms Word2Vec} 

The test perplexity of the generative modeling can achieve on the SAGE dataset with LSTM is 20.7 and with Transformer is 15.6. Transformer outperforms LSTM on generative modeling pretraining. 
We find that generative modeling as \textbf{pretrain} is sufficient for models to learn useful word embeddings and models with \textit{+Pretrain} outperforms models with \textit{+Word2Vec} on both CSU and cross-hospital dataset PP.

\paragraph{Generative modeling helps Transformer more}

In our experiment, we compare the performance of our system by adding generative modeling objective as an auxiliary task during the classification task.
Adding the generative modeling as an auxiliary task improves both Transformer and LSTM on CSU test set as well as the cross-hospital PP evaluation set. The effect of auxiliary training is more significant on Transformer than on LSTM. We also combine the generative modeling pretraining as well as the auxiliary task during the classification task and observe a substantially better performance on the overall model compared to the baseline model with either encoder. 

\section{Interpretation} 

In order to gain intuition on how deep learning models process clinical notes, we implement a gradient-based interpretation method on our model. The method attributes prediction scores to input by computing the attribution score as gradient $\times$ input~\citep{ancona2018towards}.  We compute the frequency of words that have score $\geq 0.2$ (threshold chosen to select on average 3\% words per note), use MetaMap dictionary as a filter to extract medical relevant terms, and then sort them in decreasing order. We sample 5 diseases and report the top 10 clinical relevant terms extracted by the model in the Table~\ref{tab:topwords}. Words captured by the model have high quality and agree with medical domain knowledge. Most words captured by the model are in the expert-curated dictionary from the MetaMap. Moreover, we notice that the model is capable of capturing abbreviations (i.e., `kcs'), combinations (i.e., `immune-mediated') and rare professional terms (i.e., `cryptorchid') that MetaMap fails to extract. 

\section{Conclusion}

We propose a framework that is robust for the cross-hospital generalization problem in the veterinary medicine automated coding task. By training the model on 1 million raw notes with generative modeling objective, and using state-of-the-art Transformer model, we substantially increase the performance of the framework on clinical notes annotated and gathered from a private hospital. Our framework can be applied to other medical domains that currently lack medical coding resources.

\bibliographystyle{aaai}
\bibliography{ref}

\clearpage  

\appendix 
\setcounter{figure}{0}
\renewcommand{\thefigure}{S\arabic{figure}}
\setcounter{table}{0}  % for future table handling
\renewcommand{\thetable}{S\arabic{table}}

\section{Supplementary Material}

\subsection{Model Details}

\paragraph{LSTM}

The Long short-term Memory Networks (LSTM) is a recurrent neural network with a long short-term memory cell~\citep{hochreiter1997long}. It maintains semantic gating functions specifically designed to capture long-term dependency between words. 
At time step $t$ with word embedding input $x_t$, the recurrent computation of the LSTM networks can be described in Equation~\ref{eq:lstm}. $\sigma$ is the sigmoid function $\sigma = 1 / (1 + e^{-x})$, and $\tanh$ is the hyperbolic tangent function. $\odot$ indicates the hadamard product.

\begin{equation}
\begin{split}
f_t & = \sigma(W_f x_t + V_f h_{t - 1} + b_f) \\
i_t & = \sigma(W_i x_t + V_i h_{t - 1} + b_i) \\
o_t & = \sigma(W_o x_t + V_o h_{t - 1} + b_o) \\
\tilde{c}_t & = \tanh(W_c x_t + V_c h_{t - 1} + b_c) \\
c_t & = f_t \odot c_{t - 1} + i_t \odot \tilde{c}_t \\
h_t & = o_t \odot \tanh(c_t)
\end{split}
\label{eq:lstm}
\end{equation}

\paragraph{Transformer}

Transformer was proposed by~\citet{vaswani2017attention} as a machine translation architecture. We use a multi-layer \textit{Transformer decoder} similar to the setup in ~\citet{radford2018improving}.

Let the previous layer's output as $H^{l-1} = [h_1^{l-1}, h_2^{l-1}, ..., h_T^{l-1}]$. At the first layer, these values equal to word embeddings added with a positional encoding defined in Equation~\ref{eq:pos} where $i$ indicates the dimension of the positional embedding, and $t$ indicates the position of this token in the sequence.

For the multihead attention, we first use three linear projections to transform $H^{l-1}$ to $K^{(i)}$, $V^{(i)}$, and $Q^{(i)}$ matrices. We compute the new hidden states $H^{(i)}$ according to Equation~\ref{eq:multihead-attention-full}.

\begin{equation}
\begin{split}
\mathrm{PE}(t, 2i) &= 
\sin(t / 10000^{2i/d})\\
\mathrm{PE}(t, 2i + 1) &= 
\cos(t / 10000^{2i/d})
\end{split}
\label{eq:pos}
\end{equation}

\begin{equation}
\begin{split}
\left(
\begin{array}{ll}
  K^{(i)} \\
  Q^{(i)} \\
  V^{(i)} \\
\end{array}
\right) &= \left(
\begin{array}{ll}
  W^{(i)}_k \\
  W^{(i)}_q \\
  W^{(i)}_v \\
\end{array}
\right) H^{l-1} +
\left(
\begin{array}{ll}
  b^{(i)}_k \\
  b^{(i)}_q \\
  b^{(i)}_v \\
\end{array}
\right) \\
\tilde H^{(i)} &= V^{(i)} (\mathrm{SoftMax}(\frac{Q^{(i)^T} K^{(i)}}{\sqrt{d}} \odot M) ) \\
H^{(i)} &= W^{(i)}_h \tilde H^{(i)} + b^{(i)}_h \\
\end{split}
\label{eq:multihead-attention-full}
\end{equation}

\begin{equation}
\begin{split}
\tilde H^l &= \mathrm{Concat}(H^{(1)}, H^{(2)}, ..., H^{(n)}) \\
H^l &= W_{o_2}\mathrm{ReLU}(W_{o_1} \tilde H^l + b_{o_1}) + b_{o_2} \\
\end{split}
\label{eq:transformer-block}
\end{equation}

An n-headed attention computes Equation~\ref{eq:multihead-attention-full} $n$ times and concatenate the obtained $H^{(i)} \in \reals^{T \times (d/n)}$ matrix $n$ times. In order to prevent dimension blow-up as the layer goes deeper, multi-head attention matrix $W^{(i)}_k, W^{(i)}_q, W^{(i)}_v$ all have dimensions $(d/n) \times d$. In Equation~\ref{eq:transformer-block}, we describe the \textit{transformer block}. The matrix multiplication by $W_{o_1} \in \reals^{D \times d}, W_{o_2} \in \reals^{d \times D}$ are referred to as a bottleneck computation, where $D$ is much larger than $d$.

\paragraph{Classifier}

The drawback of letting $c = h_T$ is that we are essentially reducing the information before timestamp $T$. We use a dot-product attention layer to transform $\{h_1, ..., h_T\}$ to a vector that summarizes the entire sequence $c$. The computation is defined in Equation~\ref{eq:projection-layer1}.

\begin{equation}
\begin{split}
\tilde h^{(k)}_T &= W^{(k)} h_T + b^{(k)} \\
\alpha_i^{(k)} &= \frac{\exp{h_i \cdot \tilde h^{(k)}_T}}{\sum_{j=1}^{T} \exp{h_j \cdot \tilde h^{(k)}_T}} \\
c^{(k)} &= \sum_{j=1}^{T} \alpha_i^{(k)} h_i \\
c &= \mathrm{Concat}(c^{(1)}, c^{(2)}, ..., c^{(n)})\\
\end{split}
\label{eq:projection-layer1}
\end{equation}

\paragraph{Experimental Setup}
We filter out all non-ascii characters in our documents, convert all letters to lower case, and then tokenize with NLTK~\citep{bird2004nltk}. We apply the standard BPE (Byte Pair Encoding)~\citep{sennrich2015neural} algorithm to address the out-of-vocabulary problem. BPE uses a vocabulary size of 50k. We truncate all documents to no more than 600 tokens,  padded with  start and end of sentence tokens. The word embedding dimension and encoder latent dimension are both set to 768. For the Transformer, we stack 6 transformer blocks, with 8 heads for the multi-head attention on each layer. We let the feedforward  dimension to be 2048. We implement our model in PyTorch. We use Noam Optimizer~\citep{vaswani2017attention} with 8000 warm up steps. Dropout rate is set to 0.1 during training to reduce overfitting. We split datasets into training, validation and test set (Table~\ref{table:pdtb-implicit}). All models are trained for 10 epochs. We use the validation set to select our best model and evaluate CSU test set and PP test set on our best model. We use a batch size of 10 for LSTM and a batch size of 5 for Transformer, which is the maximum allowed to train on a single GPU.

\subsection{Dataset Details}

\begin{table}[htbp]
\centering
\begin{tabular}{c|ccccc}
\toprule
 & CSU & PP & SAGE \\
 & (Labeled) & (Labeled) & (Unlabeled) \\
\midrule
\# of notes & 112,557 & 586 & 1,019,747 \\
\# of training set & 101,301(90\%) & 0(0\%) & 917,665(90\%) \\
\# of validation set & 5,628(5\%) & 0(0\%) & 51,103(5\%) \\
\# of test set & 5,628(5\%) & 586(100\%) & 50,979(5\%) \\
Avg \# of words & 368 & 253 & 72 \\
Average \# of BPE tokens & 374 & 267 & 73 \\
\bottomrule
\end{tabular}
\caption{Descriptive statistics of the three datasets.}
 \label{table:pdtb-implicit}
\end{table}

\paragraph{SNOMED-CT Codes}
SNOMED-CT is a comprehensive clinical health
terminology managed by the International Health
Terminology Standards Development Organization~\citep{donnelly2006snomed}. Annotations are applied from the SNOMED-CT veterinary extension (SNOMED-CT VET), which
is a veterinary extension of the International SNOMED-CT edition. In this work, we try to predict disease level SNOMED-CT codes. 

\begin{figure*}[htbp]
    \centering
    \includegraphics[width=5.5in]{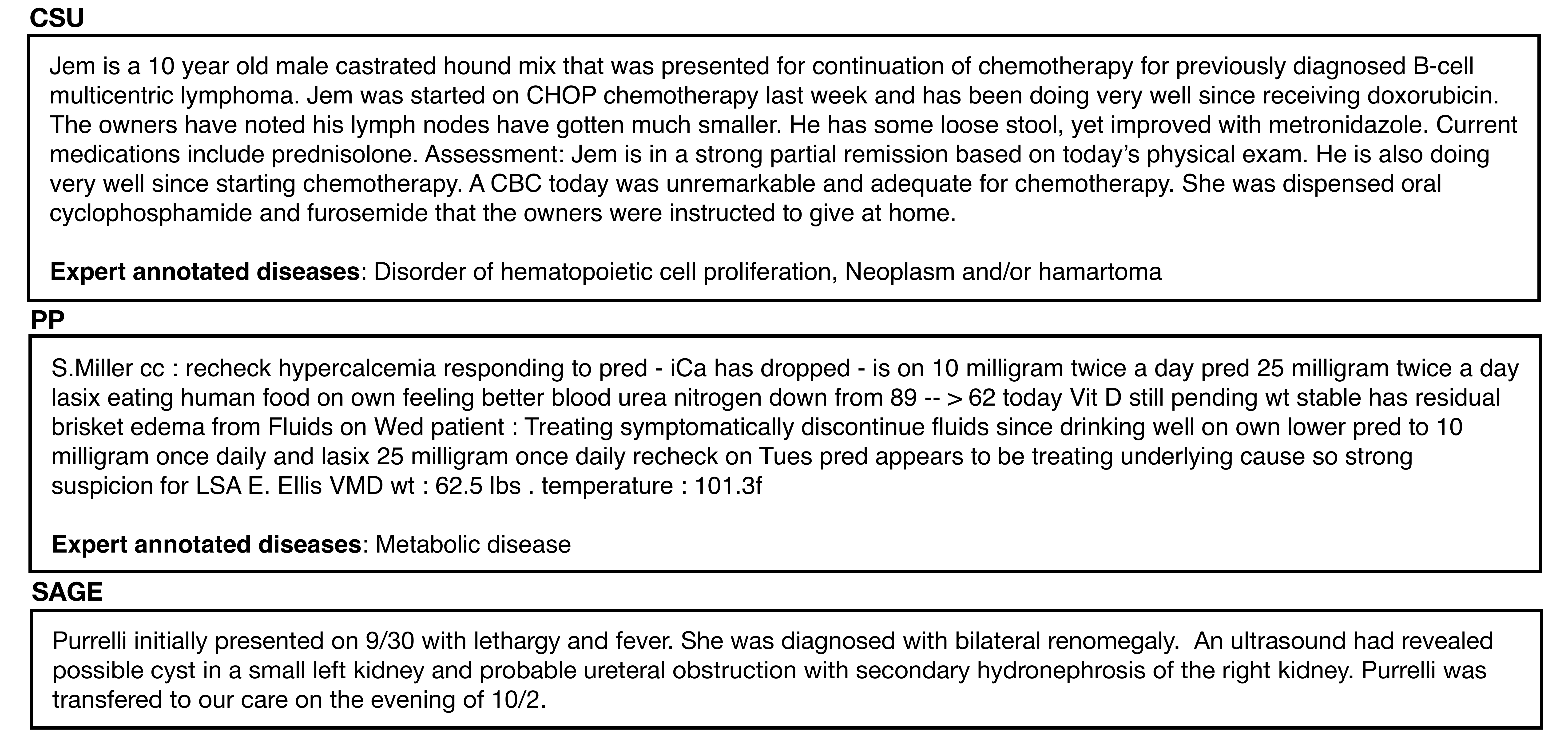}
    \caption{Examples from the CSU, PP and SAGE datasets. CSU and PP are expert labeled and SAGE is unlabeled. }
    \label{fig:example}
\end{figure*}

\paragraph{Example} We select three examples from each dataset and show them in Figure~\ref{fig:example}.

\paragraph{Length Distribution}

\begin{figure}
\begin{minipage}[!ht]{0.5\linewidth}
    \centering
    \includegraphics[width=2.5in]{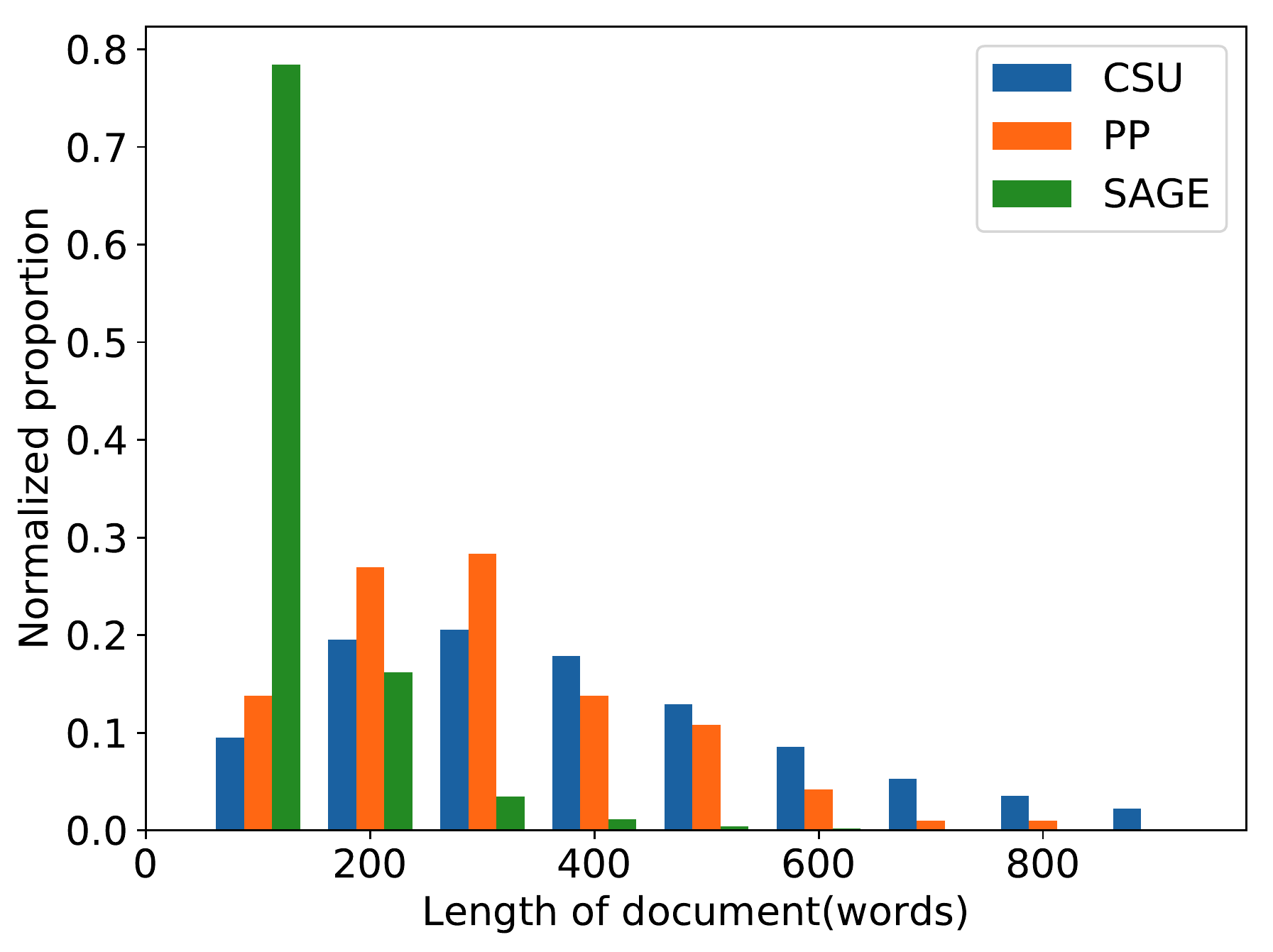}
    \caption{Document length distribution.}
    \label{fig:length}
\end{minipage}%
\begin{minipage}[!ht]{0.5\linewidth}
    \centering
    \includegraphics[width=2.5in]{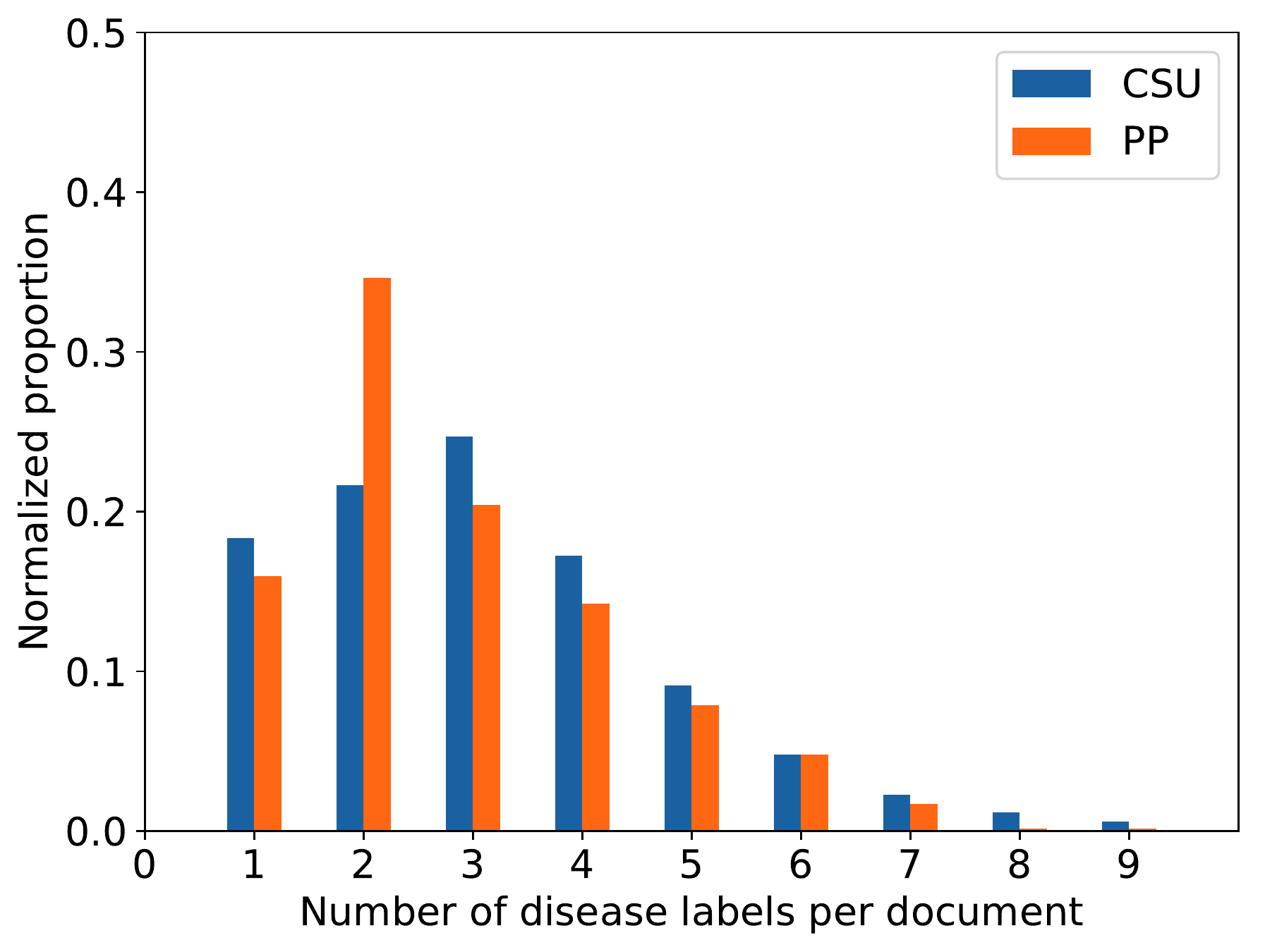}
    \caption{Label number distribution.}
    \label{fig:label}
\end{minipage}
\end{figure}

We plot a histogram to show the proportion of records in each dataset with certain length in Figure~\ref{fig:length}. 

\paragraph{Number of Label Per Document Distribution}

We plot a histogram to show the proportion of records in each labeled dataset with certain number of labels in Figure~\ref{fig:label}. 

\paragraph{Species Distribution}

\begin{figure}
\begin{minipage}[!ht]{0.5\linewidth}
    \centering
    \includegraphics[width=2.5in]{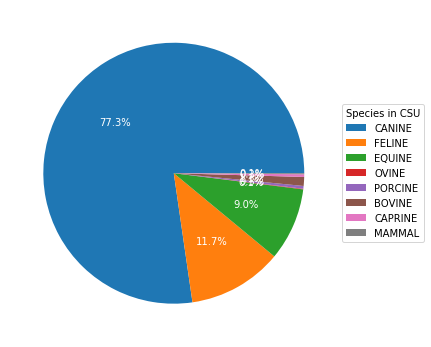}
\end{minipage}%
\begin{minipage}[!ht]{0.5\linewidth}
    \centering
    \includegraphics[width=2.718in]{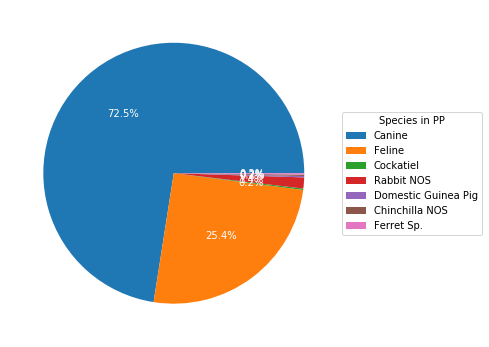}
\end{minipage}
\caption{Species distribution in CSU dataset (left) and PP dataset (right).}
\label{fig:species}
\end{figure}

We plot pie charts to show the proportion of species in each labeled dataset in Figure~\ref{fig:species}. 

\paragraph{Data Availability}

The data that support the findings of this study are available from Colorado State University College of Veterinary Medicine, a private practice veterinary hospital near San Francisco and SAGE Centers for Veterinary Specialty \& Emergency Care, but restrictions apply to the availability of these data, which were made available to Stanford for the current study, and so are not publicly available. Data are however available from the authors upon reasonable request and with permission of Colorado State University College of Veterinary Medicine, the private hospital and SAGE Centers for Veterinary Specialty \& Emergency Care.

\subsection{Result Details}
% the 20 most frequent
We compute precision, recall, F1 and accuracy score for 20 most frequent disease categories. We list the results in Table~\ref{tab:fine-grained-performance}.

\begin{table*}[!ht]
% Generated from Transformer + Pretrain + Auxiliary seed = 1234
% Allen: Add AUC if needed
\centering
\begin{tabular}{l|cccc|cccc}
\toprule
 & \multicolumn{4}{c|}{CSU} & \multicolumn{4}{c}{PP}  \\
Disease & P & R & $F_1$ & N & P & R & $F_1$ & N \\
\midrule
Disorder of cellular component of blood & 81 & 51 & 63 & 2263 & 50 & 43 & 46 & 7\\
Congenital disease & 73 & 37 & 49 & 3345 & 50 & 12 & 19 & 17\\
Propensity to adverse reactions & 75 & 81 & 78 & 5105 & 56 & 44 & 49 & 43\\
Metabolic disease & 81 & 44 & 57 & 5265 & 57 & 46 & 51 & 26\\
Disorder of auditory system & 85 & 64 & 73 & 5393 & 77 & 80 & 79 & 64\\
Hypersensitivity condition & 82 & 80 & 81 & 6871 & 50 & 44 & 47 & 50\\
Disorder of endocrine system & 81 & 70 & 75 & 7009 & 53 & 44 & 48 & 46\\
Disorder of hematopoietic cell proliferation & 94 & 90 & 92 & 7294 & 67 & 50 & 57 & 16\\
Disorder of nervous system & 80 & 62 & 70 & 7488 & 46 & 19 & 26 & 27\\
Disorder of cardiovascular system & 88 & 49 & 63 & 8733 & 91 & 19 & 31 & 53\\
Disorder of the genitourinary system & 85 & 58 & 69 & 8892 & 67 & 27 & 39 & 44\\
Traumatic AND/OR non-traumatic injury & 71 & 67 & 69 & 9027 & 52 & 58 & 55 & 19\\
Visual system disorder & 91 & 79 & 84 & 10139 & 77 & 55 & 64 & 62\\
Infectious disease & 80 & 42 & 55 & 11304 & 70 & 30 & 42 & 88\\
Disorder of respiratory system & 83 & 57 & 68 & 11322 & 47 & 26 & 33 & 27\\
Disorder of connective tissue & 87 & 56 & 68 & 17477 & 78 & 29 & 42 & 24\\
Disorder of musculoskeletal system & 87 & 61 & 72 & 20060 & 66 & 45 & 53 & 56\\
Disorder of integument & 92 & 64 & 75 & 21052 & 77 & 49 & 60 & 156\\
Disorder of digestive system & 76 & 68 & 72 & 22589 & 75 & 52 & 61 & 195\\
Neoplasm and/or hamartoma & 95 & 85 & 90 & 36108 & 43 & 63 & 51 & 59\\
\bottomrule
\end{tabular}
\caption{Performance of the best model (Transformer+Auxiliary+Pretrain) for 20 most frequent disease categories.}
\label{tab:fine-grained-performance}
\end{table*}

\begin{figure}
\begin{minipage}[!ht]{0.33\linewidth}
    \centering
    \includegraphics[width=1.7in]{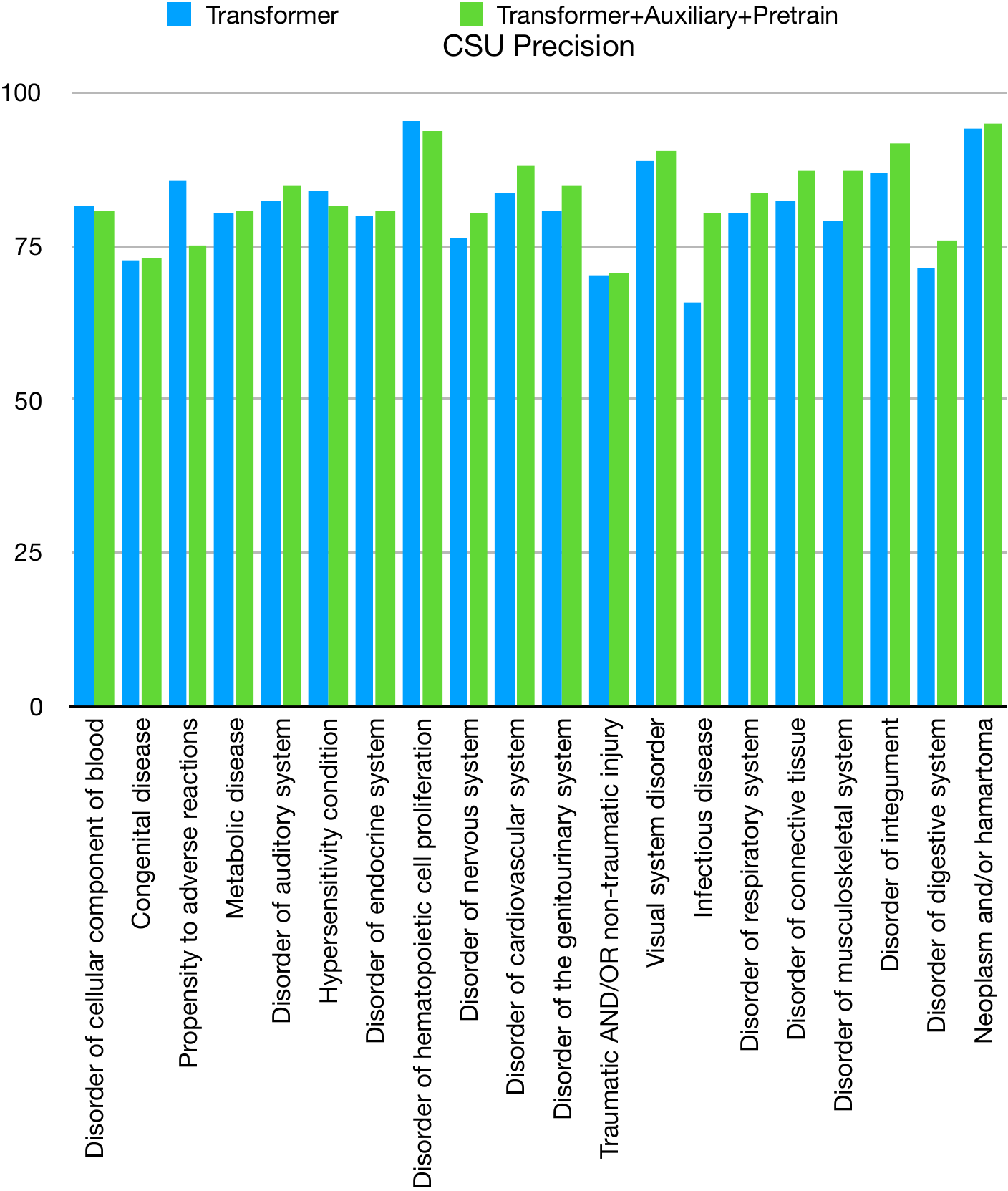}
\end{minipage}%
\begin{minipage}[!ht]{0.33\linewidth}
    \centering
    \includegraphics[width=1.7in]{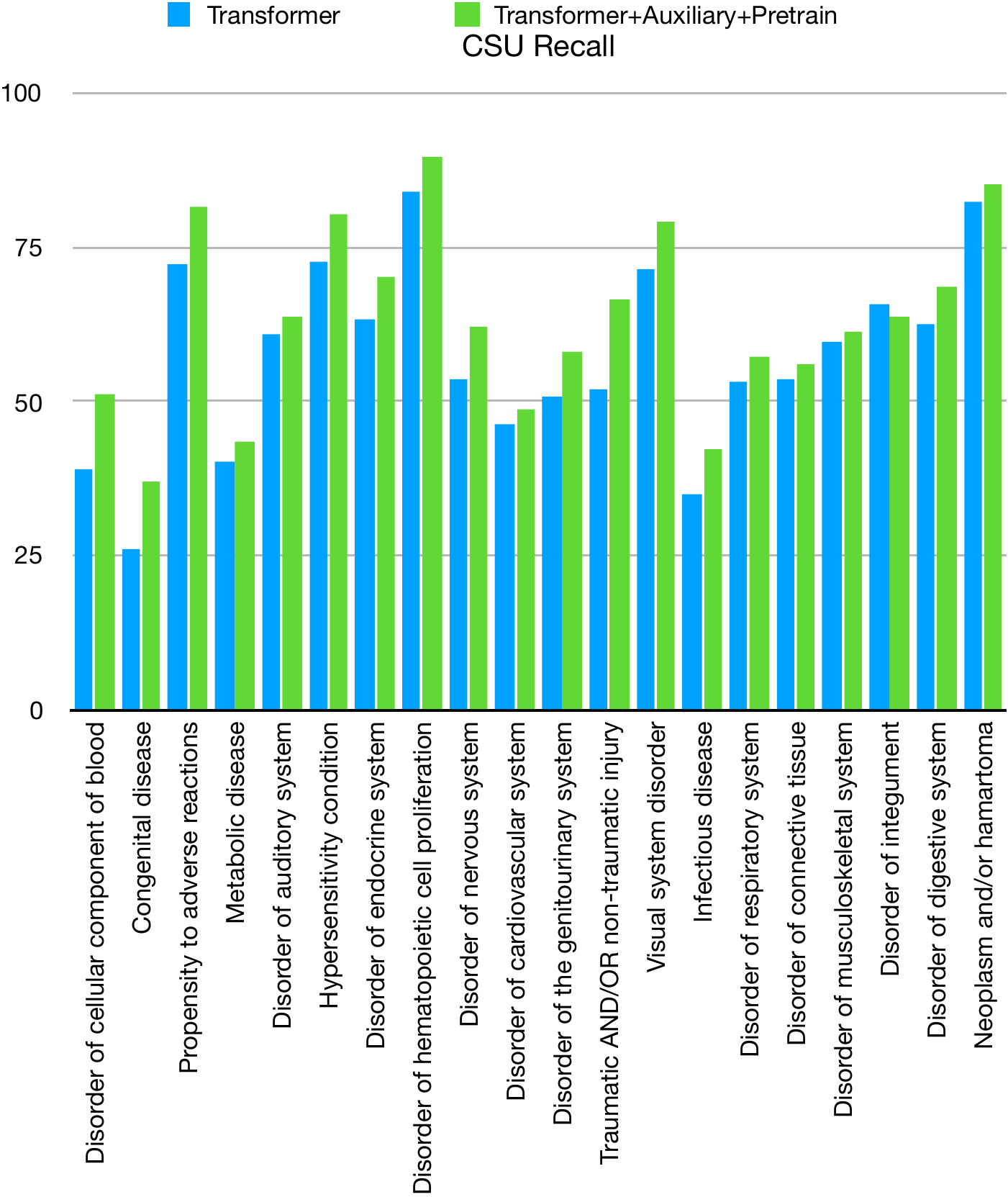}
\end{minipage}
\begin{minipage}[!ht]{0.33\linewidth}
    \centering
    \includegraphics[width=1.7in]{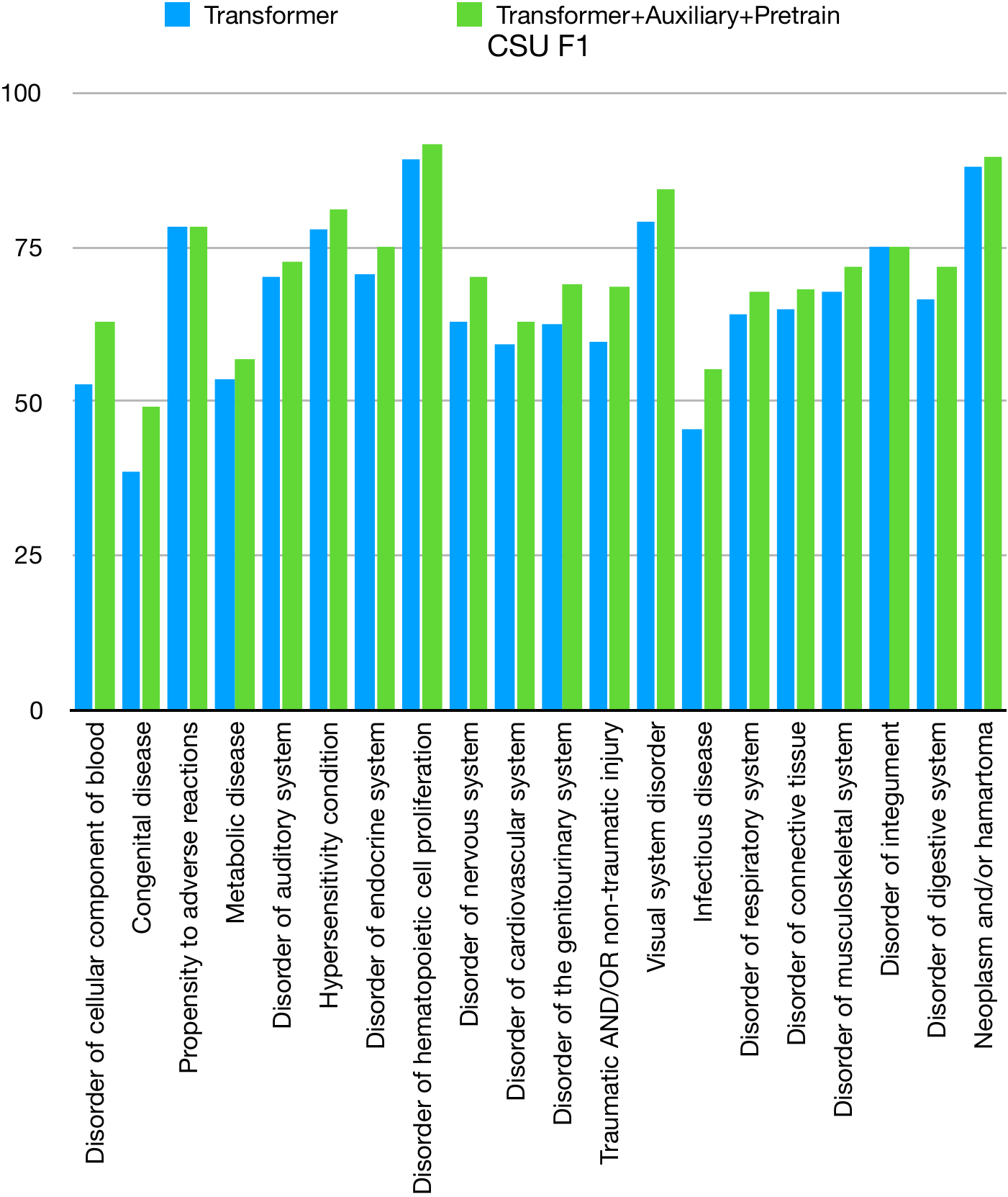}
\end{minipage}
\begin{minipage}[!ht]{0.33\linewidth}
    \centering
    \includegraphics[width=1.7in]{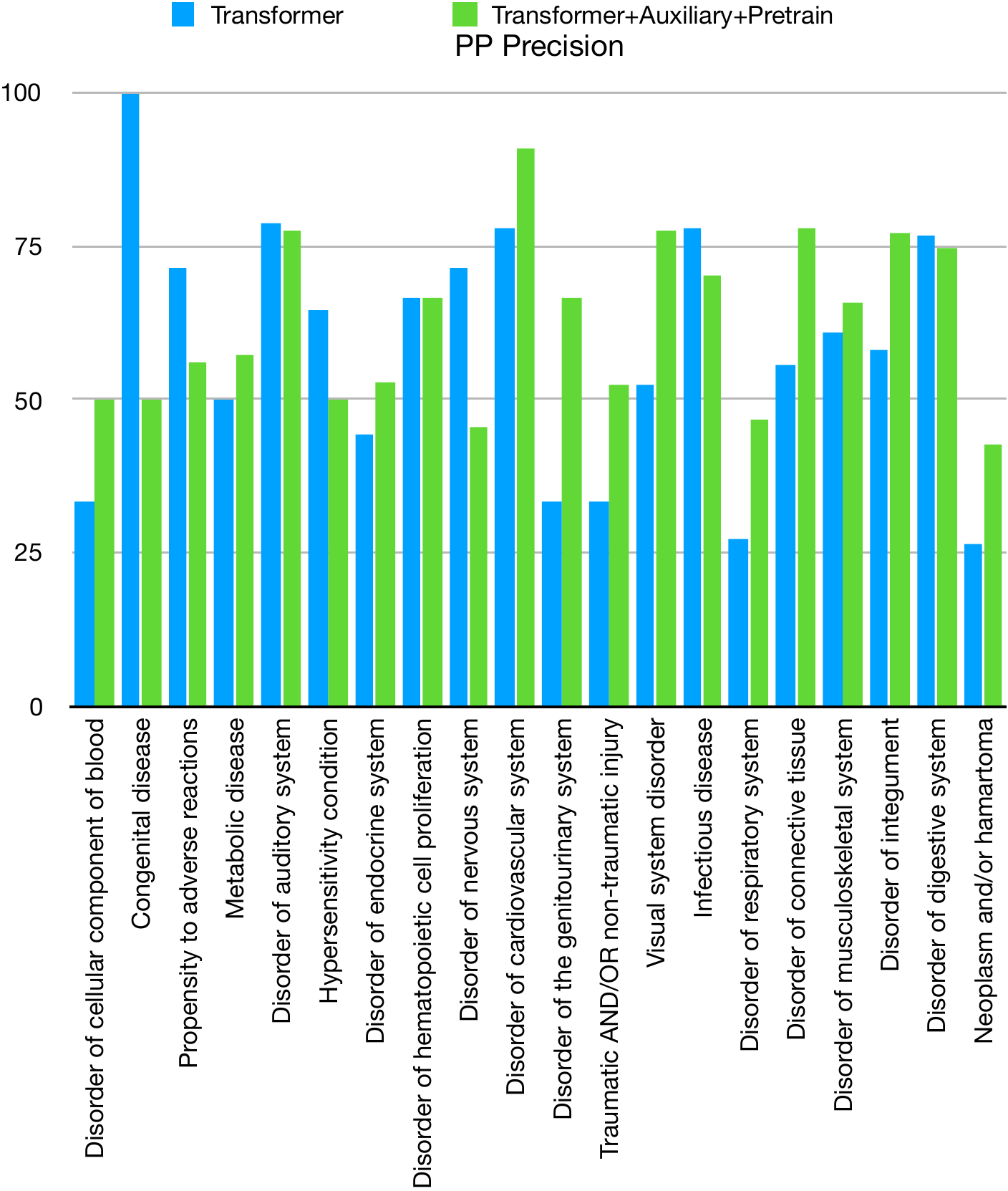}
\end{minipage}%
\begin{minipage}[!ht]{0.33\linewidth}
    \centering
    \includegraphics[width=1.7in]{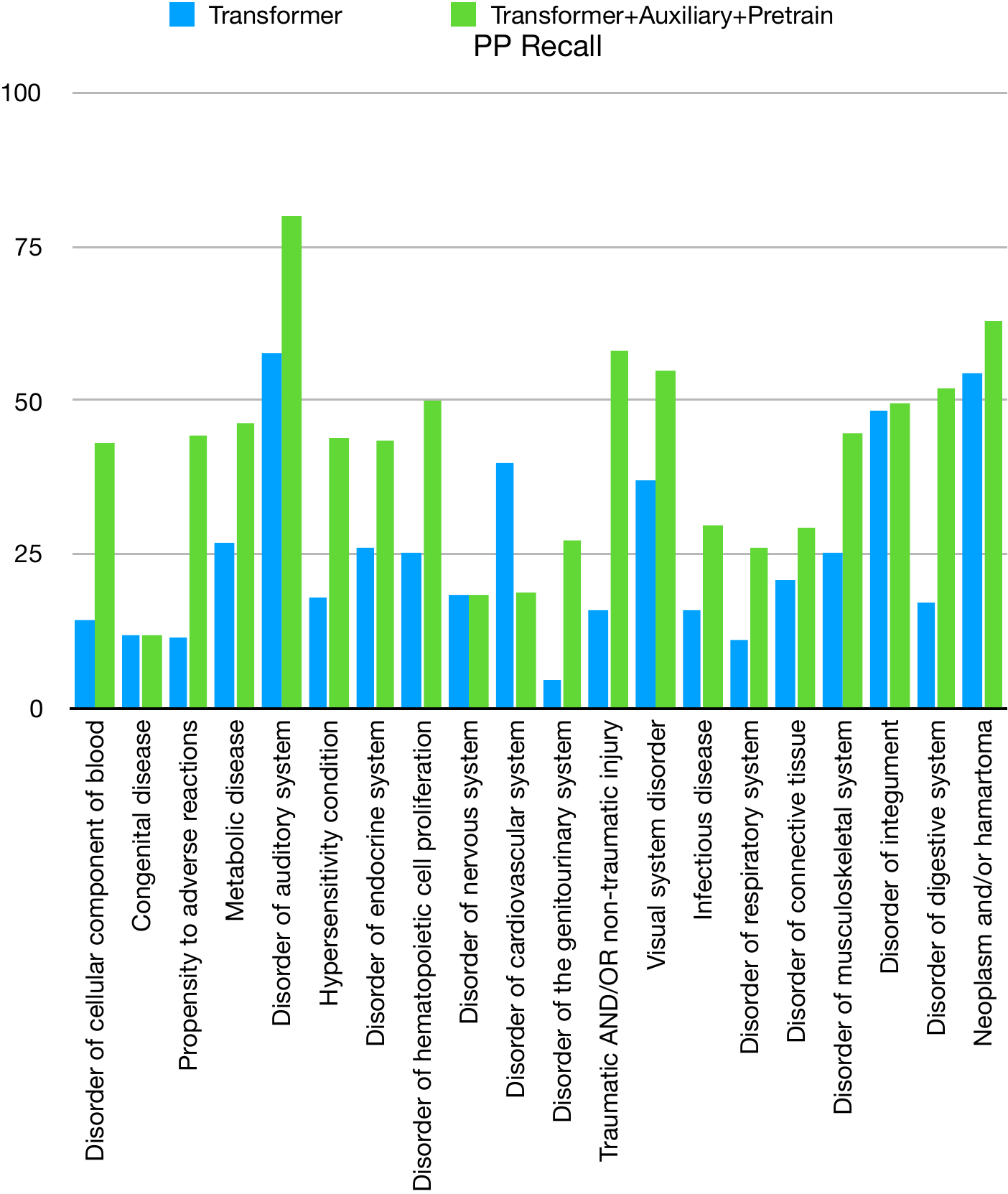}
\end{minipage}
\begin{minipage}[!ht]{0.33\linewidth}
    \centering
    \includegraphics[width=1.7in]{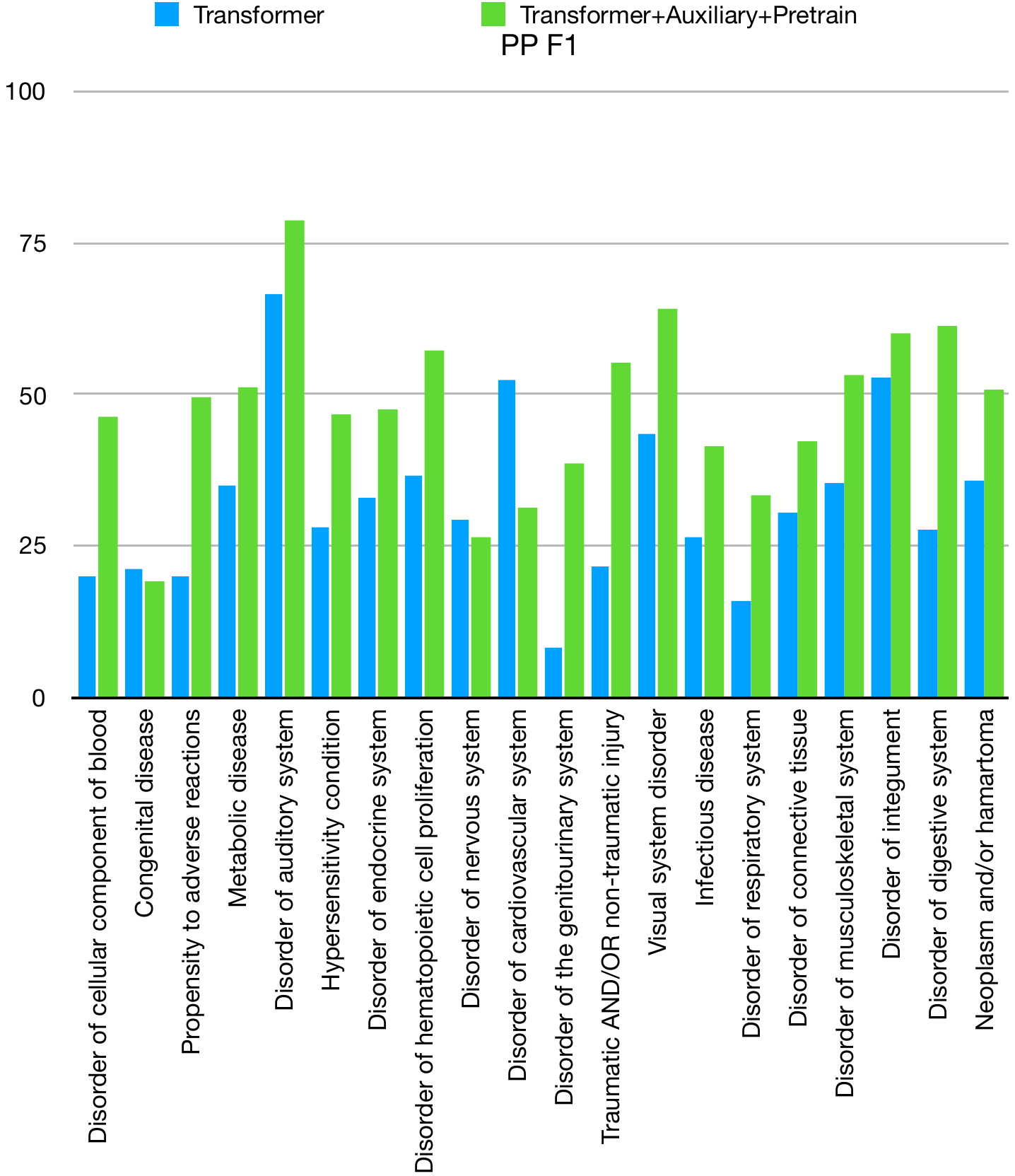}
\end{minipage}
\caption{Performance comparison on the CSU and PP dataset for the 20 most frequent disease categories. Generative modeling pretraining and generative modeling as an auxiliary task improve recall significantly.}
\label{fig:label-performance}
\end{figure}

To investigate the effectiveness of generative modeling pretraining and generative modeling as an auxiliary task, we compare the performance of two models: Transformer v.s. Transformer+Auxiliary+Pretrain on both CSU and PP datasets. We report precision, recall and F1 score for the 20 most frequent disease categories, as shown in Figure~\ref{fig:label-performance}. We observe a significant improvement in recall for Transformer+Auxiliary+Pretrain model, which explains the overall improvement in F1 score.

% \begin{figure*}[!ht]
%     \centering
%     \includegraphics[width=5.5in]{label_performance.pdf}
%     \caption{Performance comparison on the CSU and PP dataset for the 20 most frequent disease categories. generative modeling pretraining and generative modeling as an auxiliary task improve recall significantly.}
%     \label{fig:label-performance}
% \end{figure*}

% \begin{figure}[!ht]
%     \centering
%     \includegraphics[width=3in]{csu_p.pdf}
%     \caption{Precision comparison on the CSU dataset}
%     \label{fig:csu_p}
% \end{figure}
% \begin{figure}[!ht]
%     \centering
%     \includegraphics[width=3in]{csu_r.pdf}
%     \caption{Recall comparison on the CSU dataset}
%     \label{fig:csu_r}
% \end{figure}
% \begin{figure}[!ht]
%     \centering
%     \includegraphics[width=3in]{csu_f1.pdf}
%     \caption{F1 comparison on the CSU dataset}
%     \label{fig:csu_f1}
% \end{figure}
% \begin{figure}[!ht]
%     \centering
%     \includegraphics[width=3in]{pp_p.pdf}
%     \caption{Precision comparision on the PP dataset}
%     \label{fig:pp_p}
% \end{figure}
% \begin{figure}[!ht]
%     \centering
%     \includegraphics[width=3in]{pp_r.pdf}
%     \caption{Recall comparision on the PP dataset}
%     \label{fig:pp_r}
% \end{figure}
% \begin{figure}[!ht]
%     \centering
%     \includegraphics[width=3in]{pp_f1.pdf}
%     \caption{F1 comparision on the PP dataset}
%     \label{fig:pp_f1}
% \end{figure}

\subsection{Interpretation Details}

We use gradient-based interpretation attribution algorithm to compute the frequency of words that have score $\geq 0.2$ (threshold chosen heuristically), use MetaMap dictionary as a filter to extract medical relevant terms, and then sort them in decreasing order. We select the top 50 words and display words that intersect with the MetaMap expert-curated dictionary. We show results in Table~\ref{tab:full-topwords-1}, \ref{tab:full-topwords-2}, \ref{tab:full-topwords-3}. Disease categories without influential words are not shown.

\begin{table*}[!ht] 
\centering
\begin{tabular}{l|l}
\toprule
Disease & Words \\
\midrule
Disorder of  & ear, otitis, ears, therapy, yeast, \\
auditory system & allergy, assessment, infection, weeks, malassezia, \\
 & allergic, dermatology, this, disease, medications, \\
 & dermatitis, left, has, avalanche, not, \\
 & topical, drops, \\
\midrule
Disorder of  & eosinophilic, then, problem, todays, hypocalcemia, \\
immune function & cornea, dose, skin, alt, weeks, \\
 & prednisolone, not, ofloxacin, eosinophilia, old, \\
 & rhinitis, duration, currently, medicine, cam, \\
 & cephalexin, molly, pancytopenia, hyperglobulinemia, herpes, \\
\midrule
Metabolic disease & diabetes, nph, hypercalcemia, glargine, vetsulin, \\
 & weeks, home, insulin, amlodipine, dose, \\
 & dehydration, culture, eye, last, visit, \\
 & assessment, time, oncology, cll, vet, \\
 & azotemia, units, ionized, lymphoma, carprofen, \\
 & consistent, surgery, \\
\midrule
Autoimmune disease & pemphigus, itp, lupus, mycophenolate, azathioprine, \\
 & not, weeks, bear, dle, planum, \\
 & diagnosed, due, assessment, thrombocytopenia, administration, \\
 & tramadol, home, platelet, mediated, \\
\midrule
Disorder of   & lymphoma, multicentric, chop, doxorubicin, assessment, \\
hematopoietic cell & trial, continued, lsa, chemotherapy, cbc, \\
proliferation & lymph, protocol, oncology, diagnosed, treatment, \\
 & home, well, ccnu, weeks, remission, \\
\midrule
Neoplasm and/or  & oncology, lymphoma, osteosarcoma, sarcoma, mass, \\
hamartoma & home, carcinoma, assessment, metastatic, adenocarcinoma, \\
 & chemotherapy, multicentric, tumor, trial, has, \\
 & surgery, disease, time, diagnosed, carboplatin, \\
 & well, weeks, pulmonary, melanoma, treatment, \\
 & metastasis, palladia, \\
\midrule
Disorder of  & cardiology, hypertension, vasculitis, disease, current, \\
cardiovascular system & home, at, assessment, valve, amlodipine, \\
 & atenolol, infection, pimobendan, sildenafil, thrombus, \\
 & pressure, heart, blood, weeks, not, \\
 & arrhythmia, ventricular, pulmonary, internal, failure, \\
 & echocardiogram, time, iliac, hours, \\
\midrule
Infectious disease & pyoderma, assessment, infection, bacterial, therapy, \\
 & uti, urinary, culture, superficial, this, \\
 & dermatitis, treat, today, secondary, infections, \\
 & well, problem, time, urine, upper, \\
 & chloramphenicol, allergies, but, weeks, site, \\
 & home, \\
\midrule
Disorder of  & assessment, otitis, therapy, pyoderma, mct, \\
integument & vinblastine, dermatology, weeks, trial, home, \\
 & has, malassezia, ear, problem, metastatic, \\
 & allergic, this, atopic, not, eyelid, \\
 & medications, mass, \\
\midrule
Traumatic AND/OR  & fracture, wound, laceration, due, assessment, \\
non-traumatic injury & trauma, this, bandage, time, owner, \\
 & fractured, eye, surgery, fractures, she, \\
 & days, dog, may, joint, abrasion, \\
 & home, radiographs, likely, change, \\
\bottomrule
\end{tabular}
\caption{Most influential words in the best model (Transformer+Auxiliary+Pretrain). Disease categories without influential words are not shown.}
\label{tab:full-topwords-1}
\end{table*}

\begin{table*}[!ht]
\centering
\begin{tabular}{l|l}
\toprule
Disease & Words \\
\midrule
Disorder of   & thrombocytopenia, pancytopenia, itp, time, mycophenolate, \\
cellular component & count, azathioprine, prednisone, tramadol, dose, \\
of blood & hemolytic, weeks, anemia, disease, leflunomide, \\
 & steroids, eye, white, assessment, injury, \\
 & future, problem, history, cbc, \\
\midrule
Disorder of  & pneumonia, pulmonary, lung, nasal, epistaxis, \\
respiratory system & adenocarcinoma, thoracocentesis, diagnosed, rhinitis, laryngeal, \\
 & oncology, carcinoma, metastatic, paralysis, respiratory, \\
 & assessment, home, mass, upper, revealed, \\
 & necropsy, liver, consistent, chemotherapy, aspiration, \\
 & srt, this, may, pneumothorax, \\
\midrule
Vomiting & vomiting, ultrasound, chronic, assessment, findings, \\
 & scan, skin, neoplasia, hematemesis, different, \\
 & ddx, machine, nephrectomy, thickened, nodule, \\
 & somewhat, ileum, not, intestines, last, \\
 & bilateral, \\
\midrule
Disorder of  & laryngeal, seizures, his, meningioma, phenobarbital, \\
nervous system & seizure, home, signs, time, assessment, \\
 & weeks, cytarabine, myelopathy, therapy, cricket, \\
 & lesion, unremarkable, disease, hyperadrenocorticism, keppra, \\
 & paralysis, tumor, neurology, levetiracetam, diagnosed, \\
 & visit, \\
\midrule
Hypersensitivity  & dermatitis, allergic, therapy, atopic, otitis, \\
condition & pruritus, ears, assessment, allergies, dermatology, \\
 & this, infection, weeks, treatment, not, \\
 & ear, dvm, allergy, future, malassezia, \\
 & time, today, \\
\midrule
Anemia & pancytopenia, anemia, visit, hemolytic, persistent, \\
 & steroids, hypertension, neoplasia, exam, thickening, \\
 & calculi, white, inflammation, prednisolone, prednisone, \\
 & treatments, vomiting, following, not, \\
\midrule
Disorder of   & bladder, assessment, hematuria, tcc, urinary, \\
the genitourinary & urethra, mass, culture, uti, pyelonephritis, \\
system & prostatic, cystitis, ureter, chemotherapy, diagnosed, \\
 & testicle, therapy, piroxicam, disease, urine, \\
 & not, prostate, revealed, carcinoma, renal, \\
 & transitional, well, treatment, surgery, \\
\midrule
Disorder of  & thrombocytopenia, pancytopenia, itp, administration, prednisone, \\
hemostatic system & time, tramadol, bear, leflunomide, history, \\
 & service, due, count, azathioprine, hypocalcemia, \\
 & dose, mild, hypothyroidism, previous, steroids, \\
\midrule
Propensity to  & dermatitis, allergic, atopic, therapy, otitis, \\
adverse reactions & allergies, assessment, ears, infection, this, \\
 & weeks, dermatology, pruritus, dvm, not, \\
 & ear, trial, treatment, atopica, malassezia, \\
 & atopy, today, \\
\midrule
Poisoning & ingestion, assessment, toxicity, chocolate, vomiting, \\
 & charcoal, not, maya, chance, activated, \\
 & this, signs, dog, possible, rattlesnake, \\
 & time, month, monitoring, therapy, marijuana, \\
\bottomrule
\end{tabular}
\caption{Most influential words in the best model (Transformer+Auxiliary+Pretrain). Disease categories without influential words are not shown.}
\label{tab:full-topwords-2}
\end{table*}

\begin{table*}[!ht]
\centering
\begin{tabular}{l|l}
\toprule
Disease & Words \\
\midrule
Mental disorder & alopecia, screen, limb, issue, \\
\midrule
Congenital disease & dysplasia, hip, bilateral, assessment, testicle, \\
 & right, cerebellar, service, surgery, echo, \\
 & congenital, options, buffalo, mild, signs, \\
 & butternut, malformation, worse, reverse, pain, \\
 & deformity, red, elbow, management, \\
\midrule
Disorder of  & osteosarcoma, assessment, osteoarthritis, surgery, dysplasia, \\
musculoskeletal system & ligament, left, disease, carboplatin, oncology, \\
 & time, right, at, rupture, trial, \\
 & diagnosed, fracture, amputation, this, joint, \\
 & bilateral, cruciate, she, chemotherapy, tendon, \\
 & lesion, home, weeks, presented, osa, \\
\midrule
Disorder of  & methimazole, thyroid, weeks, levothyroxine, carcinoma, \\
endocrine system & mass, hyperadrenocorticism, assessment, diabetes, diagnosed, \\
 & home, disease, nph, trilostane, dose, \\
 & time, may, hyperthyroidism, surgery, visit, \\
 & glargine, eye, \\
\midrule
Disorder of  & dental, assessment, sac, adenocarcinoma, melanoma, \\
digestive system & mass, home, has, anal, time, \\
 & oncology, carboplatin, anesthesia, left, metastatic, \\
 & disease, this, enteropathy, necropsy, problem, \\
 & not, surgery, oral, lip, liver, \\
 & enteritis, from, \\
\midrule
Visual system  & eye, ophthalmology, surgery, eyelid, assessment, \\
disorder & sicca, time, uveitis, diagnosed, this, \\
 & keratitis, cataract, treatment, mass, glaucoma, \\
 & after, week, well, months, visit, \\
 & infection, \\
\midrule
Disorder of  & osteosarcoma, assessment, ligament, surgery, carboplatin, \\
connective tissue & disease, dysplasia, rupture, cruciate, fracture, \\
 & amputation, hip, weeks, right, diagnosed, \\
 & left, trial, osa, chemotherapy, anesthesia, \\
 & this, tendon, bilateral, oncology, joint, \\
 & crcl, she, well, \\
\midrule
Disorder of  & level, progesterone, high, apparently, assessment, \\
labor / delivery & draw, healthy, days, puppies, \\
\midrule
Disorder of  & progesterone, level, today, veterinary, measure, \\
pregnancy & high, labor, pregnant, approximately, assessment, \\
 & healthy, prior, once, \\
\bottomrule
\end{tabular}
\caption{Most influential words in the best model (Transformer+Auxiliary+Pretrain). Disease categories without influential words are not shown. }
\label{tab:full-topwords-3}
\end{table*}

\end{document}